%% file: root.tex
\documentclass[letterpaper, 10 pt, conference]{ieeeconf}  

\IEEEoverridecommandlockouts                              
\usepackage{siunitx}
\usepackage{hyperref}
\usepackage[]{graphicx,color}
\usepackage[utf8]{inputenc}

\input{acronyms.tex}

\title{\LARGE \bf
A GNC Architecture for Planetary Rovers with Autonomous Navigation Capabilities
}

\author{Martin Azkarate$^{1}$, Levin Gerdes$^{1}$, Luc Joudrier$^{2}$ and Carlos J.~Pérez-del-Pulgar$^{3}$
\thanks{$^{1}$Martin Azkarate and Levin Gerdes are with HE Space for ESA -- European Space Agency, 2200 AG Noordwijk, The Netherlands
        {\tt\small martin.azkarate@esa.int}}%
\thanks{$^{2}$Luc Joudrier is with the European Space Agency, 2200 AG Noordwijk, The Netherlands
        {\tt\small luc.joudrier@esa.int}}%
\thanks{$^{3}$Carlos J.~Pérez-del-Pulgar is with the Systems Engineering and Automation Deptartment, University of Malaga, 29071 Malaga, Spain
        {\tt\small carlos.perez@uma.es}}%
}

\begin{document}

\maketitle
\thispagestyle{empty}
\pagestyle{empty}

\begin{abstract}

This paper proposes a \ac{GNC} architecture for planetary rovers targeting the conditions of upcoming Mars exploration missions such as Mars 2020 and the \ac{SFR}.
The navigation requirements of these missions demand a control architecture featuring autonomous capabilities to achieve a fast and long traverse.
The proposed solution presents a two-level architecture where the efficient navigation (low) level is always active and the full navigation (upper) level is enabled according to the difficulty of the terrain.
The first level is an efficient implementation of the basic functionalities for autonomous navigation based on hazard detection,
local path replanning, and trajectory control with visual odometry.
The second level implements an adaptive SLAM algorithm that improves the relative localization,
evaluates the traversability of the terrain ahead for a more optimal path planning, and performs global (absolute) localization
that corrects the pose drift during longer traverses.
The architecture provides a solution for long range, low supervision and fast planetary exploration.
Both navigation levels have been validated on planetary analogue field test campaigns.

\end{abstract}

\section{INTRODUCTION}

\label{sec:introduction}
Autonomous navigation is considered a crucial technology to succeed on future planetary exploration missions with rovers.
In particular, \ac{SFR} could require a daily traverse of up to \SI{500}{\meter}.
Assessing an obstacle-free route of that length from ground, based only on orbital data and downloaded End-of-Sol rover imagery,
is technically improbable and practically impossible.
Autonomous navigation is therefore meant to play a key role,
since the rover must be capable of robustly localizing itself on the
Martian surface with minimum drift and of scanning the terrain ahead in order to detect hazards and
plan a safe path to its navigation target.
In this paper, we present an end to end architectural solution for the whole \ac{GNC} system
that shall drive the rover autonomously to the next exploration target while optimizing the computational 
load and allowing fast navigation speeds by adapting its architecture to the terrain conditions.

The approach we propose assumes that an orbital low-resolution map of the terrain is available and can be processed to
determine geographic characteristics of the area.
This assumption is reasonable for the case of Mars 2020 and \ac{SFR} considering 
the mapping that has already been performed for the landing site selection.
The processing of this map allows to detect big obstacles and non-navigable
areas such as mountains and craters and also to perform a preliminary terrain classification
according to different characteristics.
This terrain classification also serves for an initial assessment of the difficulty of the terrain
and for estimating its traversability.
Depending on this terrain classification and difficulty assessment,
one can decide which components or navigation functionalities should be active during each part of the traverse.
Following this concept, our \ac{GNC} architecture comprises two navigation levels such that
when the terrain is more challenging the level is increased and more functionalities are run
in order to perform a more thorough analysis of the terrain.
On the other hand, in less challenging terrain, only the low level of the architecture is active
in order to keep the navigation functionalities to a minimum, while still guaranteeing a safe autonomous traverse. 

The first navigation level, here referred to as \textit{Efficient Navigation}, is used in easy to moderate terrains.
The navigation starts with a roughly safe path provided from Ground.
The \ac{VO} module running over the images of the \ac{LocCam} keeps track of the rover's
relative localization and the trajectory control algorithm
is in charge of ensuring the rover follows the path within a certain safety corridor.
In parallel, a Hazard Detection algorithm is checking for hazards in the vicinity of the rover 
from the \ac{LocCam}'s images.
As long as no hazards are found, the rover is continuously following the provided path.
When a hazard is detected near the rover's trajectory,
a local replanning is triggered to account for the perceived obstacle.
The new path rejoins the original path as soon as the obstacle is evaded.

As soon as the rover enters a more difficult terrain,
a significant presence of hazards can actually render this navigation mode inefficient,
due to the frequent need for replanning.
In such difficult areas, the autonomy level is switched to the \textit{Full Navigation} mode,
where the \ac{NavCam} are used to map the terrain ahead and an adaptive \ac{SLAM} algorithm
based on the particle filter and scan matching methods is run.
This algorithm improves the relative localization estimate of the \ac{VO} and
produces a local elevation map that can be used to
evaluate the traversability of the terrain several meters ahead
in order to plan more optimal paths that can avoid hazards smoothly.
Finally, the local map generated by the \ac{SLAM} algorithm is used at
discrete times to perform absolute pose corrections by finding a matching correspondence
between the local and orbital maps.
This correction serves to reduce the otherwise inevitable drift in localization in the absolute reference frame.

\begin{figure*}
    \centering
    \includegraphics[width=0.9\textwidth]{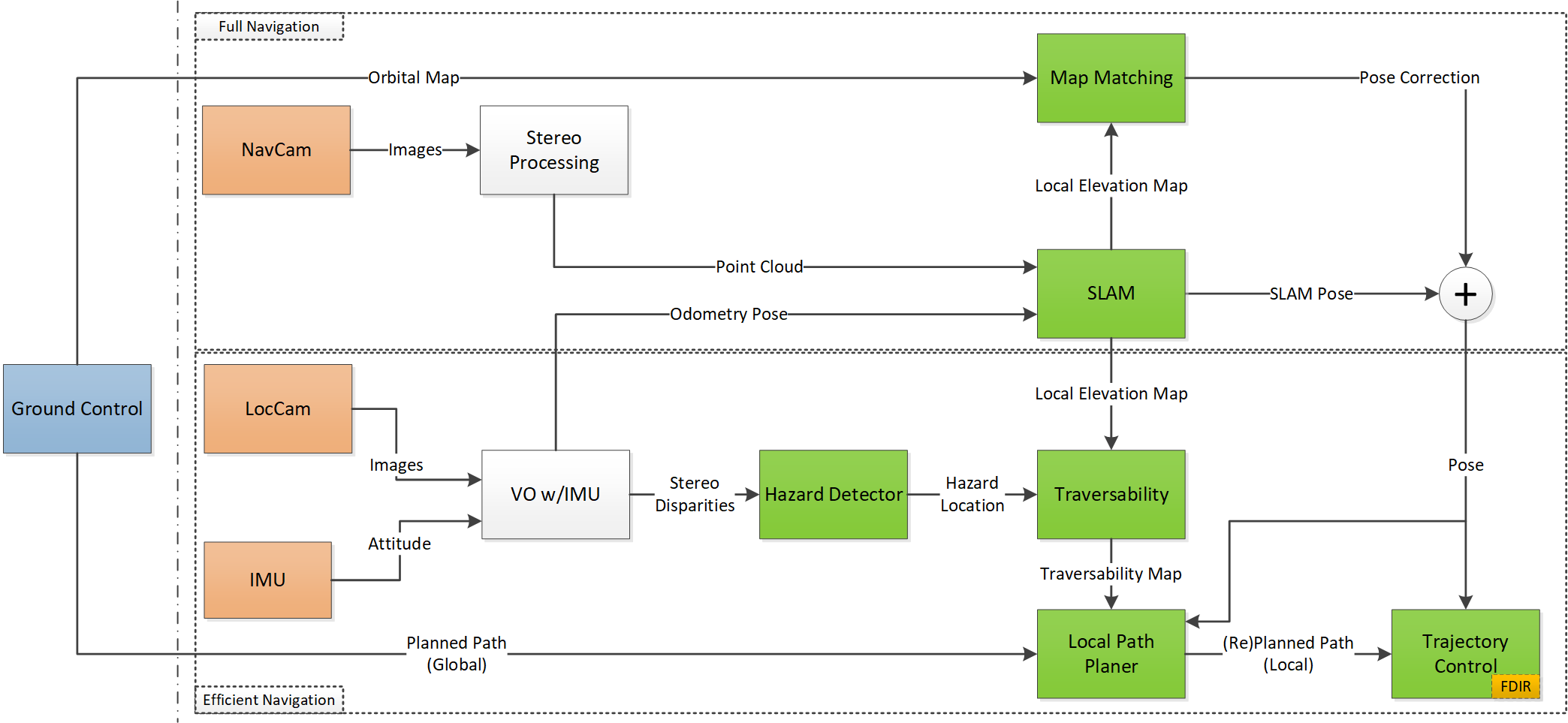}
    \caption{Schematic concept of the two-level \ac{GNC} architecture for Autonomous Navigation of Planetary Rovers}
    \label{fig:schematic}
\end{figure*}

The schematic overview in \autoref{fig:schematic} shows the conceptual design of the architecture with two navigation levels.

Regarding the paper structure, first we present in \autoref{sec:literature} a review of related work
in the context of autonomous navigation of planetary rovers.
Then, we describe the main features of the two navigation levels in
\autoref{sec:efficient-navigation} and \autoref{sec:full-navigation}.
Later, in \autoref{sec:results}, the field test results of running these two navigation levels independently are presented.
Finally, the conclusions drawn from these results and future work are described in \autoref{sec:conclusion}.

\section{RELATED WORK}
\label{sec:literature}
For the last two decades, space agencies and industrial partners have been developing  partial or full solutions for the autonomous
navigation problem, addressing the particular conditions of planetary rovers and the constraints of space systems in general.
By implementing trajectory control and relative localization using \ac{VO} \cite{Cheng2005,Shaw2013} rovers were first capable of
autonomously following a safe path provided by operators from Ground.
However, the difficulties for the operators to gauge the
traversability of the terrain at farther distances from the rover has clearly limited the traversed distance
per Sol and the overall reach of rover missions.
The need for onboard mapping and terrain analysis capabilities is widely accepted but the
resource limitations of space-grade processors have been hindering the implementation of these computationally demanding modules \cite{Lentaris2018}.
Currently, the Curiosity rover and the upcoming Rosalind Franklin (ExoMars) rover have managed to fit
autonomous navigation capabilities in their onboard processors \cite{Maimone2017,Winter2017},
however, the navigation approach they propose requires the rovers to stop for
tens of seconds every few meters in order to assess the terrain traversability and
plan a safe path for the next segment.
Other relevant works have tried to reduce the execution time of these modules by means of hardware acceleration,
i.e., porting their algorithms into FPGA solutions \cite{Kostavelis2014,Lentaris2019}.
Yet, the navigation approach for the future SFR rover remains a hard challenge 
to solve and several solutions are being considered \cite{Moreno2013,Bora2017}.
One sure thing is that the high traverse speeds required by the mission would be difficult
to achieve if the rover were to stop every few meters.

Recent works in \cite{Marc2018,Weclewski2019} have addressed the navigation
problem using a multi-mode architecture in order to achieve faster overall
traverses by adapting the computational load of the system to the terrain
difficulty.
This concept is similar to the two level architecture presented in this paper,
however the implementation of the separate navigation levels is significantly different.

\section{EFFICIENT AUTONOMOUS NAVIGATION}
\label{sec:efficient-navigation}
This navigation mode allows for fast and long autonomous traverses in terrains
of low to moderate difficulty.
Operators may have problems, even on relatively easy terrains,
to produce obstacle-free paths to be followed by the rover,
specially when the paths exceed a few tens of meters.
As already mentioned, the orbital map low resolution and the limited reach
at which operators can judge the terrain hazards from rover telemetry images
are the main issues in this case.
Therefore, we would like to provide operators with the possibility to plan less
conservatively while still ensuring the safety of the rover.
At the same time we want to leverage on the potentially higher speed that traversing not too difficult areas enables.
Because of the latter, we do not wish to engage full mapping and traversability analysis computations,
which would render it necessary to stop every few meters to compute and merge navigation maps.
Instead, we propose to run a computationally comparatively cheap hazard detection and avoidance approach,
which relies on the reuse of disparity data that is already part of the stereo odometry computations.
It is worth mentioning that in an ideal easy terrain case no hazards would be detected or needed to be avoided,
since the ground control shall aim to upload optimal safe paths.
But again, the hazard detection feature continuously ensures the rover safety,
specially in cases of hindered visibility or at farther distances.


\subsection{Overview}
\label{sec:efficient-navigation-overview}
The lower part of \autoref{fig:schematic} depicts the pipeline of the \textit{Efficient Navigation} level.
The ground control team uploads a manually planned path leading, e.g.,
from the rover's current position to the goal position of that Sol.
We refer to this path as the \textit{Global Path}.
During the execution of this path, the rover utilizes its \ac{LocCam}
and \ac{IMU} for relative localization.
This provides the estimated rover pose to the trajectory control module \cite{Filip2017}, which is responsible for driving
the rover along the path (within a certain safety corridor).
Meanwhile, the \acs{LocCam}'s stereo disparities are fed into the Hazard Detector, which analyzes them, looking for hazards in the close vicinity of the rover.
In case a hazard is detected, the rover stops,
registers the hazard in a local traversability map, and feeds this map into a local path planner.
The job of this path planner is to circumvent the hazard and rejoin the otherwise safe and planned global path as soon as possible.
Finally, a \ac{FDIR} module is connected to the trajectory control process,
implementing additional safety features such as slip ratio monitoring, rover orientation angle limits, and motor current limits.

This navigation level was already implemented in the \ac{ESA} laboratory rover prototype \ac{HDPR} \cite{Boukas2016} and tested in a planetary analogue field test campaign.
A more detailed description of the different modules that compose this level are presented in \cite{Gerdes2019} and the subsections hereafter describe its main features.

\subsection{Hazard Detection}
\label{sec:efficient-navigation-hazard-detection}
\begin{figure}
    \centering
    \includegraphics[width=0.9\columnwidth]{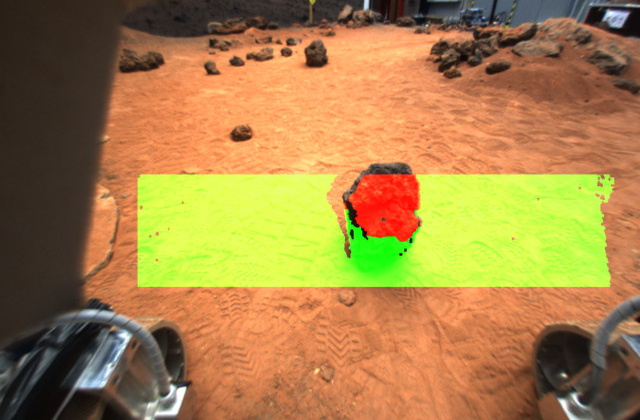}
    \caption{The hazard detector's region of interest is highlighted in green.
    The red overlay indicates a detected hazard (a big rock).}
    \label{fig:hazard_detection_roi}
\end{figure}

The key aspect for the approach to hazard detection presented here
is that it avoids performing 3D reconstructions or other operations of similar computational cost.
Instead, the algorithm simply relies on stereo image disparity values and 
precomputed calibration information.
Its main function is to take the depth information of two stereo images and just compare
their values to the calibration data in a look-up table.
Additionally, the algorithm is not run over the whole image,
but just in a \ac{RoI} right in front of the wheels
(see \autoref{fig:hazard_detection_roi}). This not only reduces the  amount of computation but is also essential for the simple obstacle detection principle, acting as an \textit{electronic bumper} protecting the rover immediate ground clearance.

To achieve a robust detection of hazards,
such as rocks that are bigger than a certain threshold,
the algorithm needs to be provided a number of static parameters.
First the position of the camera relative to the rover frame needs to be measured.
Then, the distances or depths for the pixels within the region of interest
are recorded while the rover is on a flat ground during calibration.

\newcommand{\calpoint}{\ensuremath{p_{\text{cal},i}}}
\newcommand{\caldist}{\ensuremath{d_{\text{cal},i}}}
\newcommand{\cameraheight}{\ensuremath{h_\text{cam}}}
\newcommand{\tolerancefar}{\ensuremath{T_\text{far}}}
\newcommand{\tolerancenear}{\ensuremath{T_\text{near}}}
\newcommand{\tolerateddistmin}{\ensuremath{d_{\min,i}}}
\newcommand{\tolerateddistmax}{\ensuremath{d_{\max,i}}}

Using the height of the camera \cameraheight{} as well as the nominal/calibrated distances \caldist{} 
from the camera to a flat ground,
we can compute the threshold distances for each pixel.
E.g., the minimal tolerated distance for pixel $i$ for a tolerance height of \tolerancenear{} can
be computed as
\begin{equation}
    \label{eq-toldistmin}
    \tolerateddistmin =
    \frac{\cameraheight - \tolerancenear}{\cos(\alpha)} = 
    \frac{\caldist \cdot \left(\cameraheight - \tolerancenear\right)}{\cameraheight}
\end{equation}
where $\alpha$ is the angle formed by the vertical vector from the camera to the ground and the optical vector of pixel $i$.
The values for each pixel can already be calculated as part of the calibration, so that, during the traverse,
we only need to compare the currently observed distances to these constant values
in order to detect hazards.

Once the presence of a hazards is confirmed within the camera frame,
this needs to be transformed into a traversability map within the rover frame.
This transformation can be computed using the perspective transformation from the camera frame to the rover frame.
To find a perspective transformation between two 2D frames,
four pairs of corresponding coordinates have to be provided.
Therefore, while calibrating the distances on a flat ground,
we can also record the positions of four points,
typically the corners of the region of interest, within the rover frame.

%

\subsection{Local Path Repairing}
\label{sec:efficient-navigation-local-replanning}
As soon as a hazard has been detected, the rover stops,
and the path planner receives both the rover pose and a traversability map
in the global frame.
This local traversability map is a binary representation of the rover's close vicinity, representing traversable areas
as 0 and hazards (plus a surrounding safety margin) as 1.
Based on the traversability, the planner can then replan locally in order to circumvent the detected hazards
and rejoin the avoidance path with the original, global path.
It is also possible to choose after which distance the local path should rejoin the global path the latest.
In our recent experiments we have opted to rejoin the global path as soon as possible,
because we assume that, since the global path is selected by operators on ground,
this is in principle the safest/easiest path,
while the traversability of the rest of the terrain is unknown and
potentially less favorable.
The rover may of course encounter and avoid multiple subsequent hazards,
even while performing an avoidance maneuver from a previously detected hazard.
More details about the implemented path planning algorithm and
its characteristics can be found in \cite{Sanchez2019}.

\section{FULL AUTONOMOUS NAVIGATION}
\label{sec:full-navigation}
In terrains where the likelihood of repeatedly coming across hazards is high,
the navigation mode explained in \autoref{sec:efficient-navigation}
might become less efficient, due to constant stopping and local path repairing needed to avoid the hazards.
Additionally, after replanning multiple times,
the resulting path becomes less optimal in terms of total length and traverse time.
As already explained, the global path computed on Ground should guide the rover optimally to the next navigation target while avoiding the main terrain obstacles, dead-ends and undesired areas in the global scale.
However, the presence of multiple hazards in the local scale can continuously deviate the rover from the global path
that the rover is trying to follow.
If hazards could be perceived at a distance when approaching them, a path to avoid these optimally
could be computed while still following the global path.
In principle, the further away the hazards are detected the more efficient the resulting path becomes,
but at the same time, the more perception and terrain reconstruction and analysis is required.
The navigation mode explained in this section fulfils this purpose
by performing the terrain analysis and path planning that more challenging
terrains require to traverse them safely and efficiently. 

\subsection{Overview}
\label{sec:full-overview}
The \textit{Full Navigation} level (see upper part of \autoref{fig:schematic}) activates the mapping process that allows for terrain evaluation and subsequent local path planning.
In this case, mapping and path planning are continuously running during the traverse, while in the \textit{Efficient Navigation} level the path planning is only triggered by the detection of a hazard.
The \textit{Full Navigation} also activates an additional localization process
and together with the mapping these implement a \ac{SLAM} algorithm.
\ac{SLAM} profits from the generated map to improve the pose estimation which in return helps in building a consistent map. 
Finally, the same map product of the \ac{SLAM} pipeline is used at discrete times
for map matching, i.e. the global pose correction module.
This is how the rover achieves accurate absolute localization onboard.
Both \ac{SLAM} and global pose correction provide an increased accuracy in rover localization.
This enables longer traverses without deviating from the global path which can be of key importance in future planetary missions.

The different modules that compose this navigation level are thoroughly explained
in \cite{Geromichalos2019} and briefly introduced hereafter for completeness.

\subsection{SLAM}
\label{sec:full-slam}
The approach to \ac{SLAM} presented here has been designed with focus on the Mars rover exploration case.
It targets the needs for planetary navigation and takes into consideration the particular conditions and limitations inherent to planetary exploration with rovers.
This is to guarantee that produced data is strictly needed for effective navigation and that \textit{a priori}
available information is used efficiently to minimize computation.

\begin{figure}
	\centering
    \includegraphics[width=0.9\columnwidth]{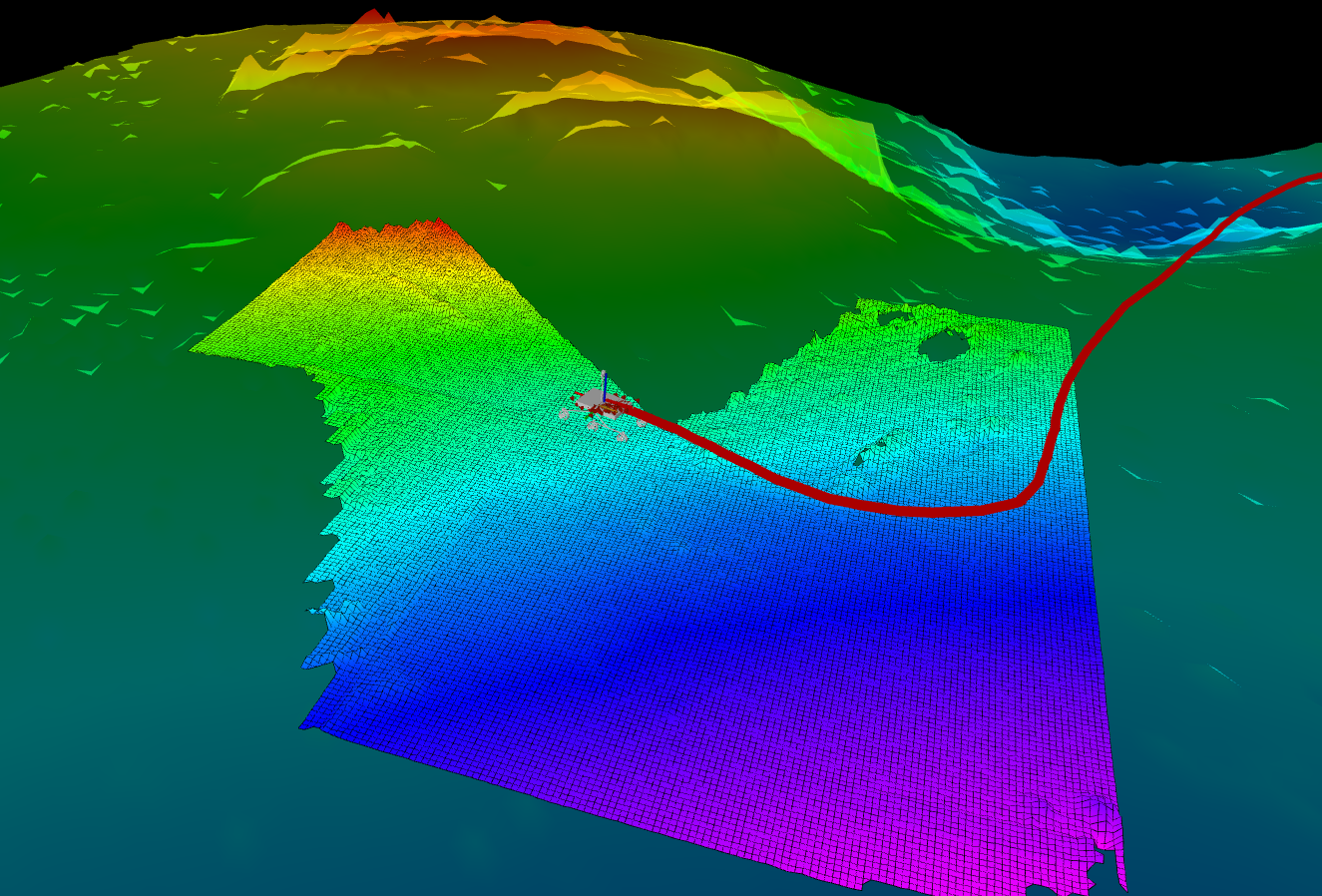}
	\caption{Perspective view of the SLAM process visualization. 
	The trajectory of the rover with respect to the initial state is represented by the red line.
	The current local elevation map and global map (background) colors represent the relative separate range of height values of each map from high (red) to low (purple).}
   \label{fig-slam}
\end{figure}

The \ac{SLAM} algorithm mainly consists of two interrelated processes:
the Data Registration (Mapping) and the Pose Estimation (Localization).
Pointclouds are used as input to the mapping process and they are pre-processed to reduce the required memory size and the amount of data to be processed by \ac{SLAM}.
Additionally, the size of the map that is kept in memory is fixed and
centered in the rover position, so that traversability information only
within the local motion range can be extracted.
We assume that, at the global scale, the rover is meant to follow a path that is computed based on orbital data.
The local map is then shifted, i.e., translated, with the input of new pose estimates and updated with each new input of pointcloud data.
The update is done by fusing two Gaussian processes following the approach of Kalman filters \cite{Cremean2005}.
The localization process implements a particle filter to estimate the pose state of the rover.
Particles are first scattered around the rover's initial position and
their state is estimated by successive prediction and update iterations.
Rover odometry (inertial, wheel, or visual) estimates are used as input for
the prediction step and input pointclouds are used for the update step.
The latter is done by scan matching the input pointcloud with the pointcloud
corresponding to the local map at the current particle state and
updating the weight of each particle in relation to the score
obtained from the scan matching.
The highest scoring particles provide the final pose estimate at each iteration.
The \ac{SLAM} process is illustrated in \autoref{fig-slam}.

\subsection{Global Pose Correction}
\label{sec:full-global}
In case of very long traverses, such as the \ac{SFR} mission scenario, even the \ac{SLAM}-based localization will eventually
drift significantly, which will potentially hinder the continuation of the exploration.
In optimal situations, Ground could compute the global localization at each end-of-sol using telemetry data and
the knowledge of the terrain.
However, the possibility of having an onboard process independent of the Ground operations is an important aspect for the mission definition and concept of operations.
Apart from reducing the dependency on communication passes,
it also virtually eliminates the limit on
traversed distance per Sol, except for the obvious energetic constraints.
The main issues with the different absolute localization techniques
developed so far (see \cite{Boukas2017,Carle2010,Chiodini2017}) is either a lack of reliability,
low precision or the need for big amounts of data (or a combination of these).
The approach presented here to estimate global pose corrections does not require big amounts of data,
it actually (re)uses the local map generated by \ac{SLAM}.
It can provide corrections that are one order of magnitude more
precise than previously known solutions and it has high reliability in terrains that are rich in elevation features, which conveniently corresponds with the type of terrains in which this navigation level is to be used.
The method consists of performing a map matching between the local map and the orbital map of the area the rover is traversing.
The local map first needs to be downsampled to match the lower resolution of the orbital map.
Then, the gradient of both maps is computed to eliminate any absolute offset in elevation between the two.
Finally, the map matching works by sliding the local map \textit{L} pixel by pixel in the orbital map \textit{O} and 
comparing the two images at each location using a metric score.
This score is stored in a result matrix \textit{R} for each location of \textit{L} over \textit{O}.
Consequently, the location in \textit{R} where the score is highest, is the location of the best match \cite{Hashemi2016}.
From the different possible metric scores we choose to compute the Cross Correlation:
\begin{equation}
R(i, j) = \sum_{i', j'} \left(L(i', j') \cdot O(i + i', j + j')\right),
\label{eq-cross_correlation}
\end{equation}
where $i'$ and $j'$ are indices describing the pixel locations in the local map whereas $i$ and $j$ are the indices of the pixels in the orbital map.


\section{RESULTS}
\label{sec:results}
In this section we present the experimental results of both navigation levels tested in separate analog campaigns.

\subsection{Experiments on Efficient Navigation}
\label{sec:results-efficient}
In July 2018, the \ac{HDPR} performed several traverses in a terrain of low to moderate difficulty that is located in the
vicinity of the \ac{ESTEC}.
This terrain is a mix of gravel, ripples, and craters suitable to test traverses
of lengths in the order of one hundred meters.
Most of the area is fairly easy to navigate, without too many slopes or obstacles.
To increase its difficulty, tens of cardboard rocks were fitted along the path.
Not only the number and density of obstacles was increased, but we also made sure
to create interesting scenarios in which more replanning will have to be done.

\begin{figure}
    \centering
    \includegraphics[width=\columnwidth]{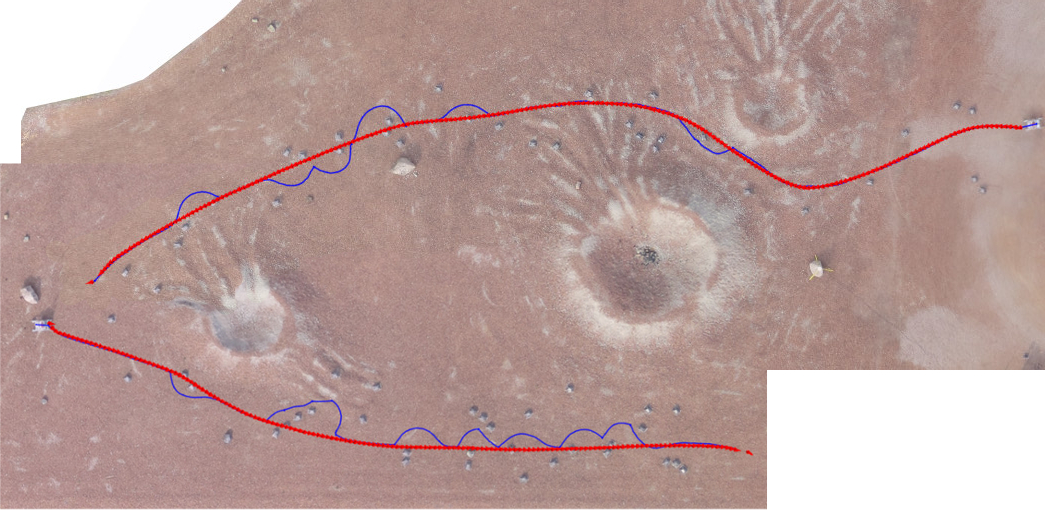}
    \caption{This figure illustrates two traverses, one in the top from the right to the left, and one in the bottom from the left to the right.
    The waypoints of the global, preplanned paths are depicted as red arrows while local path repairing is shown as blue lines.}
    \label{fig:efficient-navigation-traverses}
\end{figure}

\autoref{fig:efficient-navigation-traverses} displays two of the longer traverses.
The first traverse crosses the valley between two craters right at the beginning
and still continues far beyond that valley, following the roughly safe path prepared
on ground and trusting that the \textit{Efficient Navigation} mode will take care of
avoiding any hazards that were not visible from ground.

The second traverse takes this idea one step further in that it forces the rover to evade a lot more
obstacles during the parts of the preplanned trajectory ``hidden'' from ground operators.
This increased difficulty becomes clear when comparing the frequency of deviations from the preplanned path:
while there are 6 replannings during the first path with a preplanned length of \SI{86}{\meter},
we observe 10 replannings over the second preplanned path with a length of \SI{62.2}{\meter}.
Overall, the total traversed trajectories for both paths measured \SI{176.9}{\meter},
approximately \SI{20}{\percent} longer than the original paths with a sum of \SI{148.2}{\meter}.
Obviously, the actual length of traversed trajectories increases with the amount of replannings executed along the path.

In addition to the number of deviations, the second traverse also prominently features situations in which
the rover cannot circumvent the current obstacle and immediately return to the original path.
While avoiding one hazard, the rover encounters more hazards and has to deviate further.
The local path repairing algorithm is able to handle those situations as well, as already mentioned in \autoref{sec:efficient-navigation-local-replanning}.


\subsection{Experiments on Full Navigation}
\label{sec:results-full}
As explained in \autoref{sec:full-navigation} this mode is meant to run the \ac{SLAM} and Global Pose Correction processes
in addition to the components already running in the efficient mode.
As of the moment of writing, this complete architecture has not
been tested yet in the field, so instead, the results shown here correspond to experiments run using datasets logged in 
previous field tests in order to evaluate the performance of the \ac{SLAM} and Global Localization algorithms.
The dataset used was collected with the \ac{HDPR} at the Teide Volcano National Park of the Tenerife island in Spain during June 2017.
The collected sensor data includes optical camera image pairs from several stereo cameras with different viewpoints and baselines,
inertial data and localization ground truth data using Differential GPS.
The orbital map of the terrain was acquired with a professional surveying and mapping drone.

\begin{figure}
    \centering
    \includegraphics[width=0.9\columnwidth]{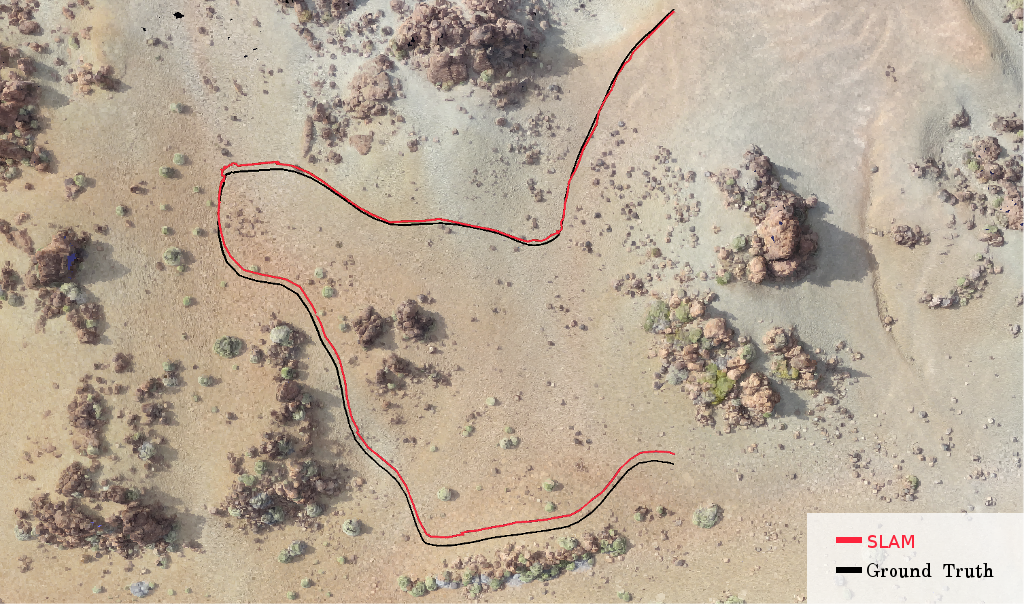}
	\caption{Orthographic views of the environment and the rover's traverses in it.
    The black and red colors correspond to the ground truth and \ac{SLAM} paths respectively.}
    \label{fig-bird_1}
\end{figure}

In the first experiment shown, the \ac{SLAM} algorithm is subject to the input of odometry data and poinclouds coming from
stereo images and its estimated localization output is compared to the traverse ground truth.
The result of the computed trajectory is shown in \autoref{fig-bird_1}.
The accumulated error of the \ac{SLAM}-based localization at the end of the trajectory is only \SI{0.8}{\%}
of the traverse distance which improves the performance of the state-of-the-art in \ac{VO} algorithms,
which typically is in the range of \SIrange[range-units=single]{1}{2}{\percent} \cite{Shaw2013,Kostavelis2016}.

In these experiments the local map generated within the \ac{SLAM} algorithm is not further processed, neither to evaluate terrain traversability nor to compute any path planning in this case,
and therefore the navigation process that follows this perception is not exercised.
However, this functionality has already been tested extensively in the efficient navigation mode
(see \autoref{sec:results-efficient}).
Additionally, the experimental results for Global Localization, shown hereafter, indirectly demonstrate the
quality of the produced local map and validate to a certain extent its suitability for navigation purposes.

\begin{figure}
    \centering
    \includegraphics[width=0.9\columnwidth]{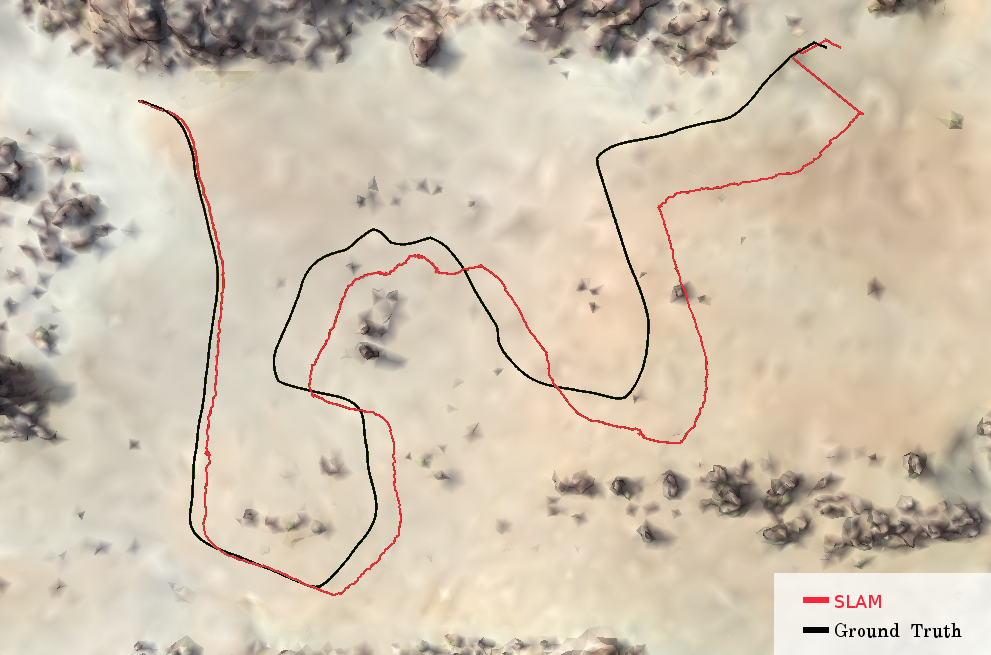}
	\caption{Orthographic views of the environment and the rover's traverses in it.
    The black and red colors correspond to the ground truth and \ac{SLAM} paths accordingly.
    The pose correction is visible in the end of the traverse.}
    \label{fig-bird_2}
\end{figure}

As explained in \autoref{sec:full-global}, the onboard generated \ac{SLAM} local map is used at discrete times
to perform a global pose correction.
The objective of this test is to examine the presented approach for absolute localization and
evaluate the achieved accuracy.
For this reason, we first run the \ac{SLAM} algorithm in a long-range dataset in order to let 
the localization estimate drift and build up relative error.
Finally, we execute the map matching process once the traversed distance has passed a certain threshold and
when the local map presents certain feature characteristics (non-flat terrain).
As shown in \autoref{fig-bird_2},
the map matching technique successfully manages to correlate the maps and apply the pose correction
that brings the pose estimation error down to values smaller than the global map resolution ($<$\SI{0.5}{\meter}).

\section{CONCLUSION}
\label{sec:conclusion}
The presented results demonstrate the benefits of both navigation modes and
their suitability for terrains of different difficulty.

The efficient navigation mode can be used in easy to moderate areas where
a roughly safe global path computed on ground can be followed.
This mode allows the rover to effectively reach a navigation target autonomously with low computational overhead
while performing local repairs on the path when hazards are encountered.
It is worth noting that the efficient navigation mode can run virtually without stopping, as it only requires to stop
shortly for local path repairing when a hazard is detected.
However, this navigation mode does not account for the localization error that inevitably builds up along the traverse.
Therefore, this mode should have margins considered in the global path if it were to be employed in a long traverse.

The full navigation mode is suitable for more difficult terrains where the global path will likely encounter several
hazards and a more optimal local path planning is needed to avoid them.
The additional benefit of the full navigation mode is the improved accuracy in localization:
firstly, in relative terms reducing the drift along the traverse, and
secondly, in absolute terms by calculating global pose corrections.
This makes the mode suitable for longer traverses where the drift in localization can increase considerably.

In conclusion, a \ac{GNC} architecture that combines the use of both modes is proposed for future planetary rover missions 
such as \ac{SFR} in order to meet the difficult requirements on fast long traverse.
While the full navigation is needed for a reduced localization drift in long traverses the efficient navigation can speed up 
the traverse in terrains that allow for it. It is also important to highlight that the efficient hazard detection can run 
continuously as an independent function in parallel to the obstacle avoidance of the full navigation, providing an interesting level
of robustness for missions where navigation is critical.
The selection of the mode can be made on Ground based on orbital imagery and terrain
classification or dynamically on board depending on the effective traverse speed and difficulties encountered on the terrain.
A field test campaign to exploit the combined use of both modes is planned to be
run by the end of 2019 where relevant metrics to assess their performance shall be measured.

\addtolength{\textheight}{-12cm}   

\bibliographystyle{IEEEtran}
\bibliography{IEEEabrv,IEEEexample}

\end{document}

%% file: acronyms.tex
\usepackage{acro}
\newcommand{\newac}[2]{\DeclareAcronym{#1}{short=#1,long=#2}}

\newac{CNES}    {French National Centre for Space Studies}
\newac{CV}      {Computer Vision}
\newac{DEM}     {Digital Elevation Map}
\newac{DSI}     {digital servo inclinometer}
\newac{ESA}     {European Space Agency}
\newac{ESTEC}   {European Space TEchnology and Research Centre}
\newac{FMM}     {Fast Marching Method}
\newac{FoV}     {field of view}
\newac{GA SLAM} {Globally Adaptive SLAM}
\newac{GNC}     {Guidance{,} Navigation{,} and Control}
\newac{GNSS}    {Global Navigation Satellite System}
\newac{HDPR}    {Heavy Duty Planetary Rover}
\newac{ICP}     {Iterative Closest Point}
\newac{IMU}     {Inertial Measurement Unit}
\newac{LocCam}  {Localization Cameras}
\newac{MER}     {Mars Exploration Rovers}
\newac{MRO}     {Mars Reconnaissance Orbiter}
\newac{MSL}     {Mars Science Laboratory}
\newac{NavCam}  {Navigation Cameras}
\newac{NASA}    {National Aeronautics and Space Administration}
\newac{PRL}     {Planetary Robotics Lab}
\newac{PTU}     {Pan-and-Tilt Unit}
\newac{PanCam}  {Panoramic Cameras}
\newac{RAMS}    {Reliability{,} Availability{,} Maintainability{,} and Safety}
\newac{ROCC}    {Rover Operations Control Centre}
\newac{RoI}     {region of interest}
\newac{RTK}     {Real Time Kinematic}
\newac{SFR}     {Sample Fetching Rover}
\newac{SLAM}    {Simultaneous Localization and Mapping}
\newac{TGO}     {Trace Gas Orbiter}
\newac{TRL}     {Technology Readiness Level}
\newac{ToF}     {time-of-flight}
\newac{VO}      {Visual Odometry}
\newac{FDIR}    {Fault Detection Isolation and Recovery}